# Analysis of Unstructured High-Density Crowded Scenes for Crowd Monitoring

Alexandre Matov[1, †]

[1] DataSet Analysis LLC, 155 Jackson St, San Francisco, CA 94111

[†] Corresponding author:

email: matov@datasetanalysis.com

## ABSTRACT

**Introduction:** Ensuring public safety in crowded areas is an important task. In recent years, with the widespread deployment of surveillance equipment and the rapid development of computer vision algorithms, real-time detection and tracking of potentially threatening, organized crowd movements from surveillance video data to ensure public safety has attracted great attention. Despite significant advances in the detection of crowded scenes based on surveillance video data over the past few years, challenges remain in terms of real time, accuracy, and automation.

**Methods:** We propose a new approach to address this problem by creating a software system that combines traditional target-tracking algorithms with classification and generative models in deep learning for real-time, accurate, and automated threat detection and aversion. To this end, we are developing an automated system for detection of complex organized movements in human crowds. Our computer vision algorithm uses network flow theory and bi-objective minimization. It can extract information from videos of crowded scenes and automatically detect and track groups of individuals undergoing organized motion that represents an anomalous behavior in the context of conflict aversion.

**Results:** Our system can detect organized cohorts against the background of randomly moving features and we can estimate the number of participants in an organized cohort, the speed and direction of motion in real time, within three to four video frames, which is less than one second from the onset of motion captured on a CCTV. We have performed preliminary analysis in this context of pedestrian and vehicle road traffic, and biological cell data containing up to four thousand features per frame and will extend this numerically to a hundred-fold for public safety applications.

**Conclusions:** Our approach shows great potential in preventing emergencies, helping to reduce social losses, and improving public safety.

## INTRODUCTION

Ensuring public safety in crowded areas, such as airport terminals and concert halls, is a daunting task (Backhouse, 1985; Fruin, 1993; Tikkanen, 1989). Issues associated with aggressive crowd behavior can occur within a few seconds with devastating consequences (Helbing et al., 2007). Because of the high density of people in such situations, it is time consuming for the security personnel to instantly realize (i) where a group of people is coming from, (ii) how many are in the group, and (iii) how long would it take for them to reach a sensitive area, such as an airport gate. For a computer vision algorithm (Haghani and Lovreglio, 2022), the estimation of these three inquiries takes an instant. Our approach to feature motion tracking will only report those feature that move together, that is, in an organized fashion forming spatiotemporal patterns, even if they are not adjacent to each other and are separated in space. This means that if several people standing far away from each other suddenly start running at the same time toward a common point of interest, the software would instantly report their number, exact locations, direction and speed of motion.

Information regarding the emergence of pockets of confrontation in human crowds is crucial for minimizing dangerous situations. Even the most vigilant human observer would not be able to instantly alert of a developing conflict. CCTV cameras (Bruch, 1942) are widely used at public venues both indoors and outdoors. The software system we propose can be installed and run, with low memory requirements, in real time. It requires a tenth of a second to detect any organized motion of multiple people. It can also dissect the motion of crowds when they intersect. If provided with a list of sensitive locations in an area, such as the coordinates of building entrances or other gates, it would generate an automated alert in real time, before any human observer would be able to see the developing situation. Such a system will notify multiple organizations that are responsible for public order and distribute a detailed set of metrics, such as number of people, direction and speed of motion, and expected time of arriving at particular sensitive places. Utilizing a computer vision-based automated system for crowd monitoring will ensure higher confidence in the response of law enforcement to a variety of conflict situations by providing, in great detail, a quantitative breakdown of the enfolding situation and, thus, allowing for a suitable response. It will also minimize the delay of response and, this way, lower the chance of undesirable consequences.

## RESULTS

## INSTANTANEOUS FLOW TRACKING ALGORITHM (IFTA)

### Maximum flow, minimum cost

Recently, we proposed a new method for instantaneous flow tracking, which we abbreviate as IFTA, based on graph theory that can be used to establish correspondence between moving features in time-lapse image sequences (Matov et al., 2011). The strength of our algorithm is the ability to detect spatiotemporal patterns in organized cohorts of features within a larger cohort with unstructured feature motion. We build triplets based on a search area for each feature, that is, the maximal displacement computation is limited based on a priori information of the video analyzed (that is, a human can run no faster than 44.7 km/h) and utilized a cost function modified from the approach Sethi proposed (Sethi and Jain, 1987), and computed the Mahalanobis distance (Mahalanobis, 1936). Each detected feature is initially linked in multiple candidate triples – in the case of the mitotic spindle, where the feature density is high (Vid. 1), this can be up to 50 triplets (Fig. 1). We then use linear programming (Murty, 1985) optimization to select the links that maximize the flow through the graph while minimizing the cost of linking by solving the primal simplex method (Dantzig, 1987). We benchmarked the performance of the algorithm with synthetic data with very high density in which the features participated in up to 120 triplets and the specificity was 96% (Matov et al., 2011).

**Interdigitating flows**

There exist prior algorithms for the tracking of coherent motion (Ali and Shah, 2007; Ali and Shah, 2008; Brostow and Cipolla, 2006; Couzin and Krause, 2003; Moussaid et al., 2009; Rabaud and Belongie, 2006; Zhou et al., 2012) in the context of the analysis of multi-directional and sometimes anti-parallel flows, which, however, are not interdigitated. The difficulty in solving the motion patterns when the features of the different cohorts interdigitate and intermingle is that the distance traveled by the features between two video frames can oftentimes be significantly higher than the average distance to their nearest neighbor (Matov et al., 2011). Such scenarios (Matov, 2024b) often lead to wrong, false positive assignments by the established tracking methods. For these reasons, we employ local and global motion models, which are parameters-free and data-driven, and a bi-objective optimization procedure to ensure that links from one time-frame to the next are established only when there is evidence for organized motion within one of the anti-parallel cohorts. We rely on a Bayesian statistics approach to fine-tune the weights of the objective function and the solution converges (<5% of the links change) in three to four iterations without requiring user parameter input.

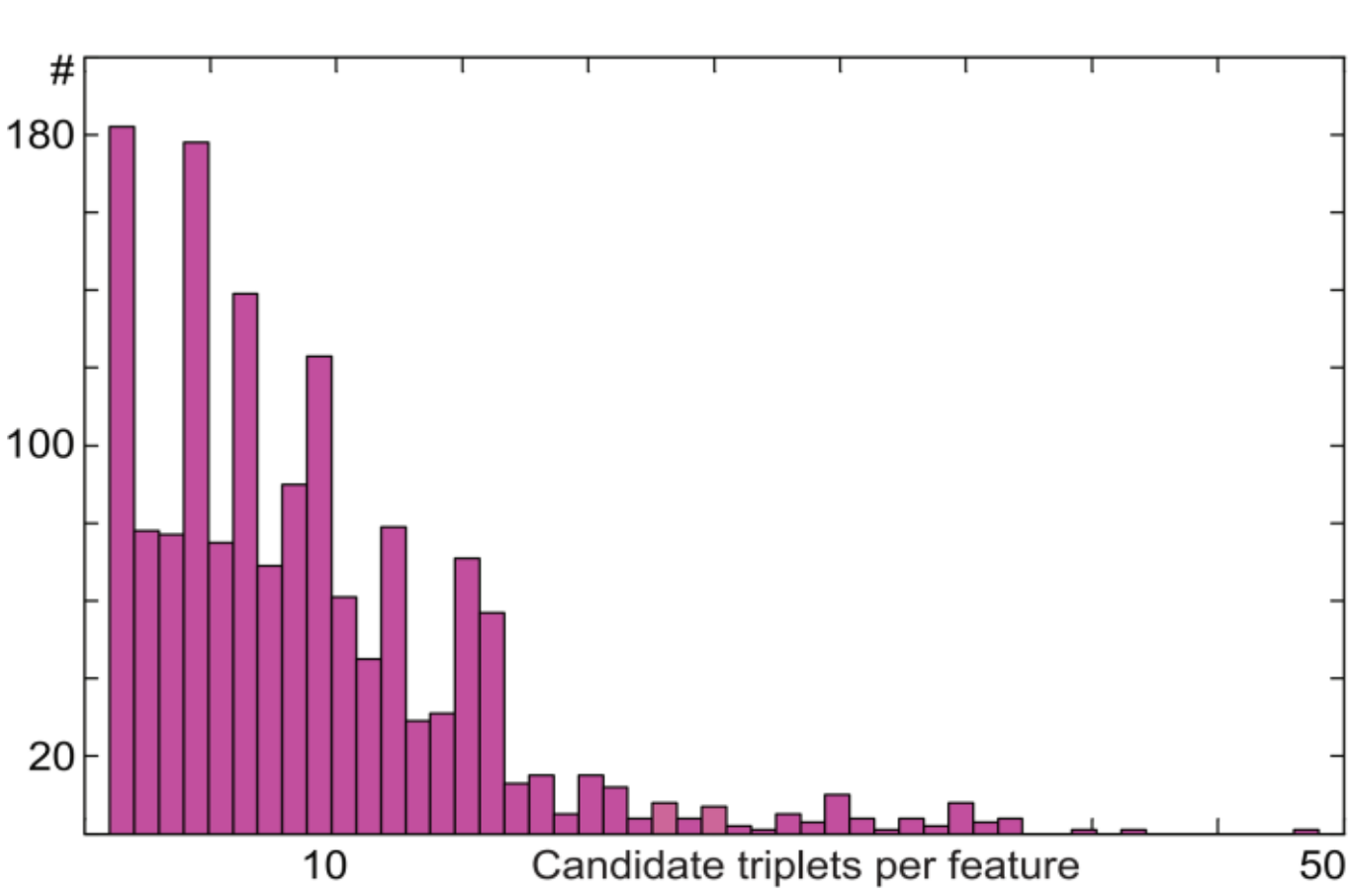

**Figure 1. Number of triplets each feature is tentatively linked to.** The network flow algorithm of IFTA establishes multiple candidate triplet links for each image feature. The optimal solution is computed via cost minimization. Because of the high density of features in the mitotic spindle in Vid. 1, some features participate in up to 50 candidate triplets.

To implement the angular components of our motion models, a distinct feature of our approach is that it always considers three time-frames at once (Matov et al., 2011; Vallotton et al., 2003), that is, we always analyze triplets of images. This allows us to study the distribution of the tentative angular motion of individual features in comparison to the overall directions of motion of the identified cohorts. In this context, our cost function for linking features is built of three elements, which penalizes links that are not equidistant in the steps from frame 1 to frame 2 and from frame 2 to frame 3, also links that display motion with very sharp turns as well as links that do not conform with the overall motion direction of the overall cohort. Following a circular *k*-means clustering (Matov et al., 2011), we apply spatial regularization and the vectors belonging to each of the cohorts (or flows) are filtered with an anisotropic Gaussian (Weierstrass, 1885) distribution with a large standard deviation along the flow direction ($\sigma_1$) and a small standard deviation perpendicular to the flow direction ($\sigma_2$). A typical ratio $\sigma_1/\sigma_2$ that we use in our analysis is 4, for instance ($\sigma_1$, $\sigma_2$) = (12, 3) or (16, 4), depending on the application. Our global motion model is calculated based on the angle each vector forms with the associated flow vector, that is, it reflects how much a vector is affected by the directional smoothing. This way, we inherently favor linking features that move together in spatiotemporal patterns, including accelerating and/or curve patterns, and can be assigned with a minimal penalty to one of the cohorts (Matov et al., 2011).

Video 1 shows a time-lapse video of cell division and a highly dynamics mitotic spindle, which – in patient cells – is hallmark of cancer (Matov, 2024c; Matov, 2024d). In these images, the microtubules (MTs) – elongated polymers building the mitotic spindle (Mitchison and Salmon, 2001) – are fluorescently labeled and appear with a punctuated pattern (Waterman-Storer and Salmon, 1998). As the structure slightly moves, image registration was done based on cross-correlation analysis during which the translation was corrected via evaluating the covariance matrix by computing the convolution of image pairs (Mitchison et al., 2004). In the video, it is visible the anti-parallel flows of the fiduciary polymer markers.

Figure 2 shows a comparison between the tracking results of IFTA (with its raw links (Fig. 2A) and the clustered and directionally filtered links (Fig. 2B)) and the feature point tracking algorithm developed in (Sbalzarini and Koumoutsakos, 2005) (Fig. 2C-D) and based on concepts borrowed from the decay of turbulence in a rotating environment (Dalziel, 1992). While this algorithm can track diffusion, it faces difficulties in resolving organized motion. The blue circle in Fig. 2C highlights the area of erroneous links, which then results in forming links across adjacent MTs laterally, while the motion is vertical. The features moving upwards are displayed with black arrows and the features moving downwards are displayed in red arrows. In the lower part of Fig. 2D, the black arrows erroneously form links between features laying on neighboring MT polymers - probably due to the high features density coupled with the lack of a motion model applied in the method.

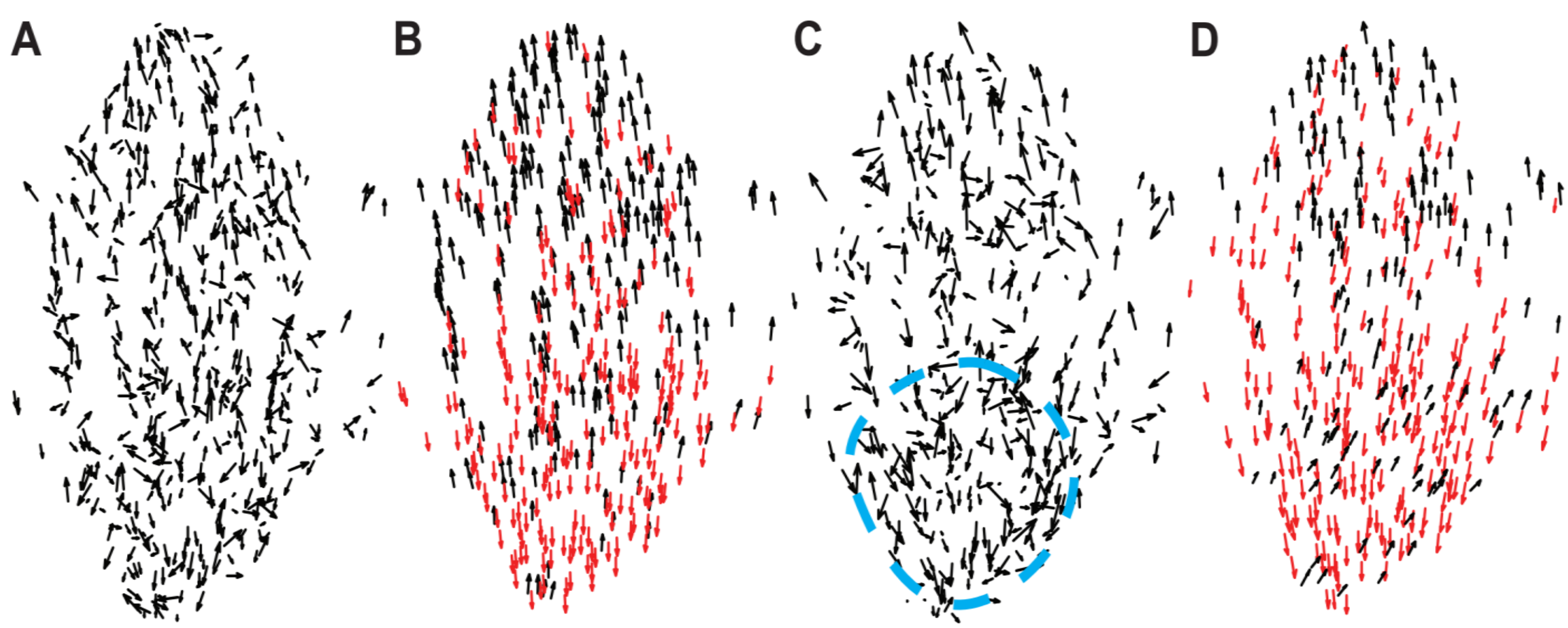

**Figure 2. Comparison of tracking results of bi-directional MT polymer flux in a metaphase mitotic spindle of *Xenopus laevis* egg extracts.** Two vertical anti-parallel flows (Vid. 1) are tracked in these images. The population of feature moving upwards is displayed in black arrows, the population moving downwards – in red arrows. (A) and (B) Tracking results obtained with IFTA. (A) Raw vectors. (B) Vectors after spatial regularization. (C) and (D) Tracking results obtained with the point correspondence algorithm developed in (Sbalzarini and Koumoutsakos, 2005). (C) Raw vectors. The blue circle highlights the area of erroneous links. (D) Vectors after spatial regularization. Based on the erroneous raw vectors, this step results in forming lateral links across polymers – since there is a very high level of proximity of adjacent MTs, that is, the speckles are closer in lateral direction – wrongly indicating movement to the right in the lower part of the spindle (black vectors).

One of our motion models (the global motion model), penalizes the deviation in terms of angular orientation of the raw vectors with the estimated overall flow. In the example of the mitotic spindle (Vid. 2), it is calculated based on the data shown in Fig. 2A (the raw vectors) and Fig. 2B (computed flow fields). The two flow fields are obtained by clustering with *k*-means (Steinhaus, 1957) of the raw vectors, which allows to assign two different populations of features – moving upwards and moving downwards (Fig. 3).

**Optimal flow, minimum cost**

Most tracking algorithms maximize the number of links (Ford and Fulkerson, 1962; Goldberg and Tarjan, 1988; Goldberg and Tarjan, 1990; Sedgewick, 1992; Shafique and Shah, 2005; Vallotton et al., 2003; Vallotton et al., 2017; Veenman et al., 2001), thus introducing false positive links, for example, when tracking cohorts of features that do not have a constant number of participants from one time-frame to the next, because of occlusions. To improve on that strategy, we introduced graph pruning. In this context, the most important feature of our approach is the application of bi-objective optimization. We use Pareto efficiency (Pareto, 1897), a method widely used in microeconomics, which Vilfredo Pareto conceived in the 19th century after realizing that two concurrent aims could be competing by definition and the most optimal solution is the point on the convex curve, which is closest to zero, the so-called utopia point. This way, we compute the optimal point of a trade-off between forming as many links as possible on one hand, which inherently increases the overall cost (or penalty), and minimizing the cost of linking on another. That is to say that by posing competing aims to our solver, our objective is to only select near-perfect links.

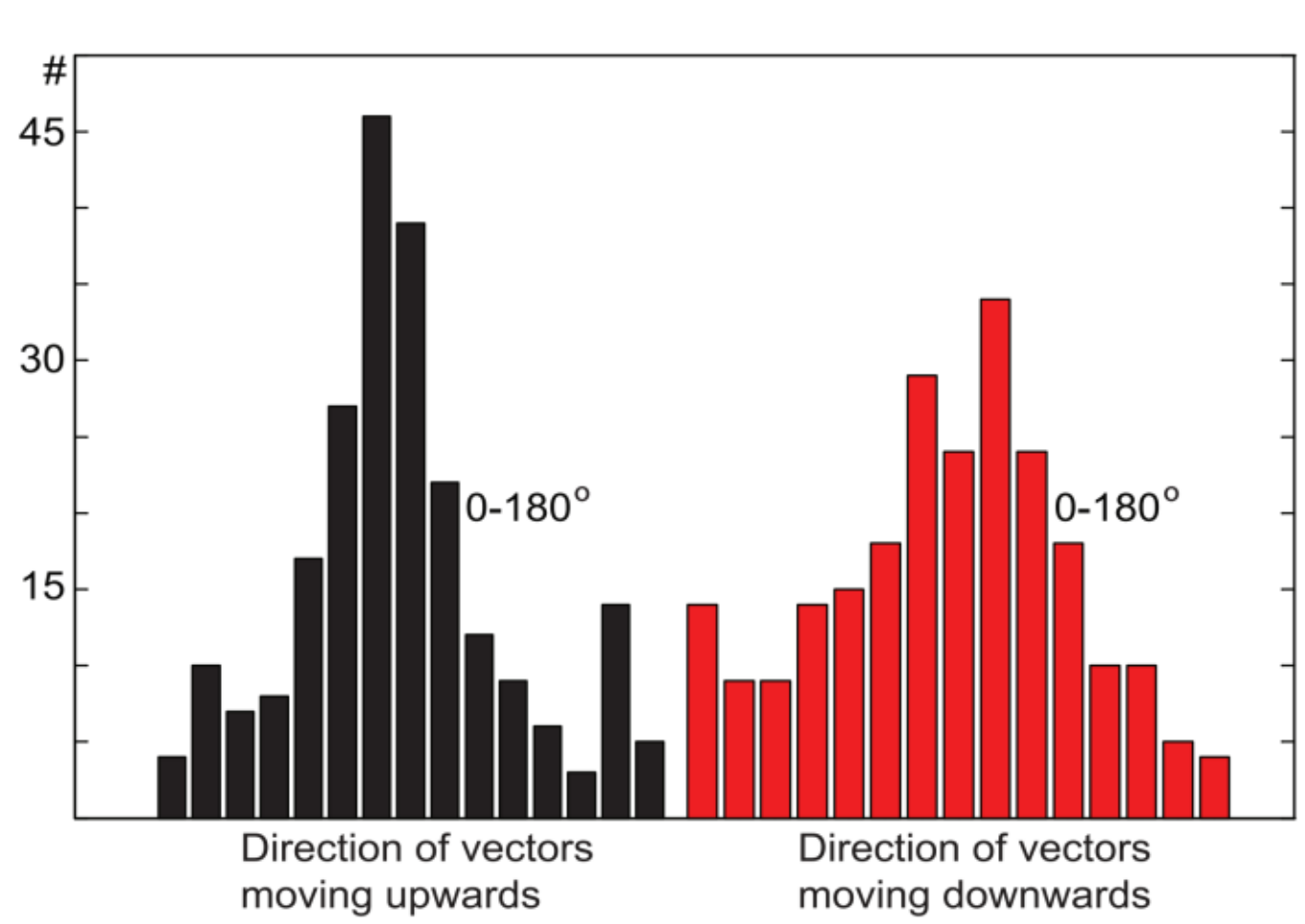

**Figure 3. Angular orientation of the raw vectors in each of the directions.** The vectors are clustered and the resulting assignment for each of the directions is shown in black for features moving upwards and in red for features moving downwards.

This way, we reduce to an absolute minimum the false positive selections we make, which is of critical importance. The weights of the Pareto optimality multi-objective function are based on Bayesian statistics, which makes the algorithm self-adaptive with rapid convergence within several iterations (most commonly within three to four iterations less than 1% of the links change between selection steps). During the analysis of the mitotic spindle shown in Fig. 1, the optimization of the cost function weights converged after five iterations, when only 0.5% of the triplets changed in comparison to the previous iteration (Fig. 4).

We evaluated the performance of another tracking algorithm, which uses three different motion priors – models of Brownian motion and directed movement with constant speed or acceleration – using a maximum likelihood association (Genovesio et al., 2004; Genovesio et al., 2006). The software can track a total number of 20,000 trajectories and the image sequence contains about 50,000 features, which form short trajectories and new features continuously appear through the video. This represents a tracking challenge and the software does not track a significant proportion of the features.

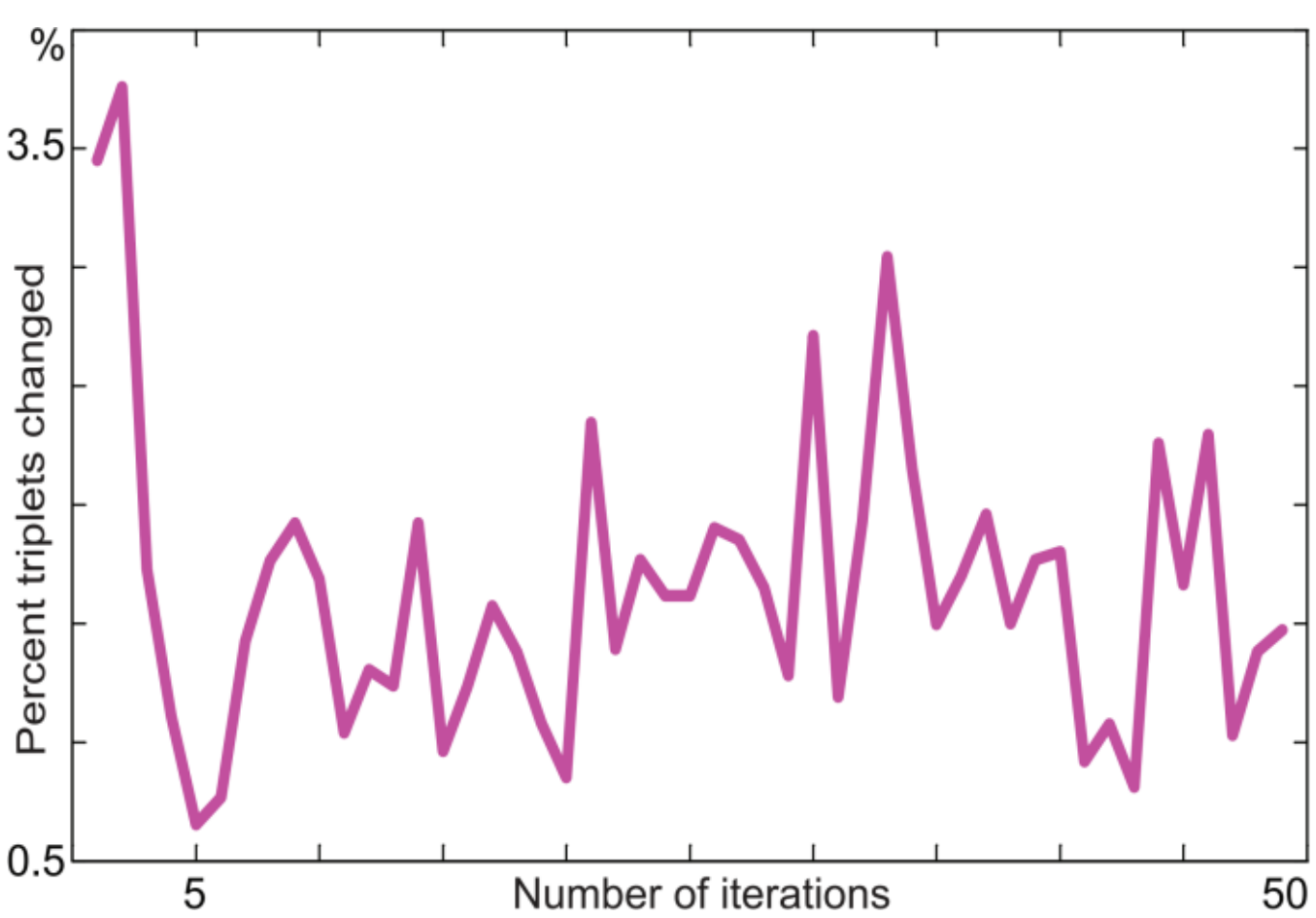

**Figure 4. Percent triplets changed as a function of the number of computational iterations.** After five iterations, only 0.5% of the triplets formed change. If we increase the number of iterations, the number fluctuates, but never exceeds 3%.

In addition, the trajectories often represent nearest neighbor tracking and link features laterally (Fig. 5), which is erroneous, as all features move vertically in one direction only and form spatiotemporal patterns, that is, either roughly half of the features move only upwards and the other half move only downwards (Vid. 1, Fig. 2B).

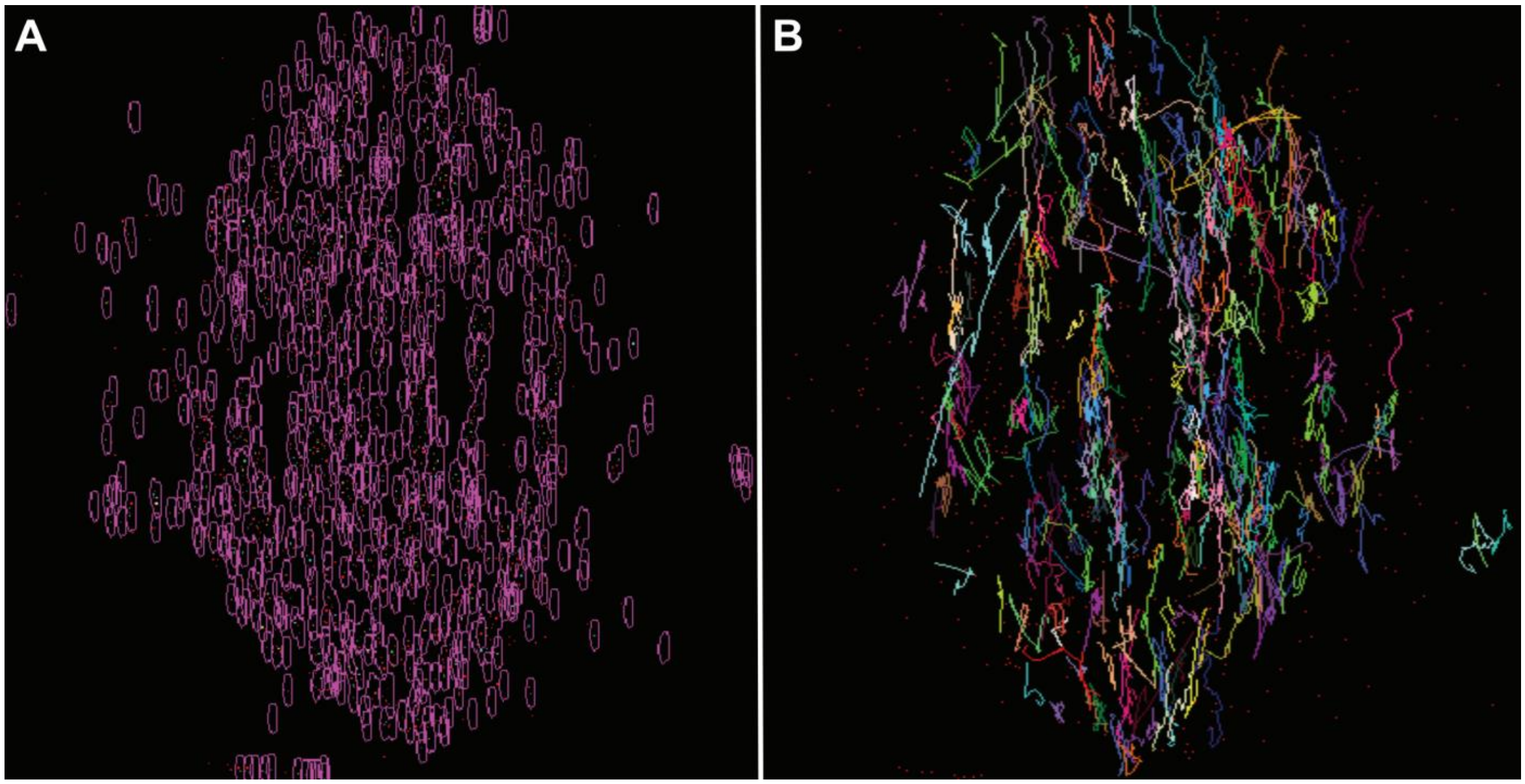

**Figure 5. Tracking results of bi-directional MT polymer flux in a metaphase mitotic spindle of *Xenopus laevis* egg extracts by the point correspondence algorithm developed in (Genovesio et al., 2006).** (A) Feature detection was done with IFTA to avoid any bias caused by using different detectors. (B) Trajectories are displayed in different colors. Many features remain untracked (see the red dots) and those tracked often seem exhibit lateral motion, which is erroneous (see Vid. 1).

## Vector clustering on a circle

In the simple scenario of two cluster centers from where vectors originate in each direction radially, we developed and implemented a circular *k*-means algorithm (Matov et al., 2011), which allows us to assign each vector to a cluster and then re-calculate the cost values based on the updated flow field. In scenarios when we are not sure of the number of overlapped flow fields, we utilized a maximization-optimization algorithm (Figueiredo and Jain, 2002; Ueda and Nakano, 1998) to identify the number of cohorts moving in different directions at the same time (we perform tests for the existence of between one and nine flows). For instance, if several gates at an airport terminal are boarding passengers at the same time and there are several groups of people moving toward each of the gates and these groups intersect in the middle of the terminal, we will track the exact number, location, speed and direction of motion for each moving individual and will assign the passenger to the corresponding cohort. The algorithm is not limited in the number of cohorts it can detect and track. To compute circular expectation-maximization, the assignment uses a mixture of Tikhonov/von Mises distributions (Bouberima et al., 2010).

**Scale space and feature selection**

Scale space theory was popularized during the 1990s through the work of Tony Lindeberg (Lindeberg, 1993). It explores, through a cascade of different levels of Gaussian smoothing (Weierstrass, 1885) of an image, the changes in perception depending on the scale of observation. Our initial feature detection kernel, in the context of biological cell signals, was a difference of Gaussians (DoG) bandpass filter (Matov et al., 2011; Wilson and Giese, 1977). The upper frequency cut-off, reflecting the limit of the optical transfer function (Fourier, 1822) beyond which the collected signal is white noise, was computationally fitted to the Bessel function forming the point-spread function of the imaging system (Bernoulli, 1728), that is, matched to the optical resolution limit of the collection frequency spectrum based on the dye emission wavelength and the diffraction limit of the objective. The lower limit relates to the computational clipping of image background, that is, non-specific signal forming larger shapes in the body of the biological cell and the cut-off was empirically obtained from the available imaging datasets to minimize the false positive feature selections. The ratio of the standard deviation of the two 2D Gaussian functions we obtained was 1.1 (Matov et al., 2011). David Lowe performed a similar analysis for the scale-invariant feature transform (SIFT) and obtained a ratio value of about 1.4 (Lowe, 1999).

The DoG detection approach allowed us to select significant protein features with high precision at a very low computational cost (Matov et al., 2010; Matov et al., 2011), which will make it suitable for detecting humans at a long distance, when people appear as single moving dots. When the images contain features that are not diffraction-limited in size, such as the processing of human crowds at a short distance, the feature selection can be based on detectors such as SIFT (Lowe, 1999; Matov, 2025), speeded-up robust features (SURF) (Bay et al., 2008; Matov, 2024g), or (particularly important for real-time analysis) ORB (Rublee et al., 2011). ORB includes modified for faster analysis versions of (oriented) features from accelerated segment test (FAST) (Rosten and Drummond, 2006) and (rotated) binary robust independent elementary features (BRIEF) (Calonder et al., 2012). The strength of these algorithms is that they detect features that persist along scales of observation and can be reliably tracked in time, even in scenarios with different lighting and partial occlusion. Having multiple features detected for the same object will also facilitate our flow tracking step since it will generate several vectors with a very similar cost. In the case of tracking motion in human crowds, the bag-of-bags-of-words model (Ren et al., 2013) can be applied to the analysis of individual images in order to add an additional component of the cost function. It offers an improvement to the bag-of-words model, because of the inclusion of spatial information during classification.

**Sliding window flow vectors and Markov Random Field**

Because our algorithm uses a global motion model that relies on directional averaging of vectors in order to compute the cost to assigning a link based on the vector immediate environment (Geusebroek et al., 2003), when the images are sparse and in a certain area there are not many features to interpolate over, the flow estimation fails. To circumvent this potential issue, in the original implementation we utilized a "sliding window" approach in which vectors from five image triplets (from time-frames 1-2-3, 2-3-4, 3-4-5, 4-5-6, and 5-6-7) were aggregated at each time step. This approach always generated the correct flow direction results, but it slowed down the performance of the tracker. To improve on this strategy, we have added the utilization of Markov Random Field (MRF), which runs as an iterative process over the solution for adjacent flow fields, thus modifying the selection results to reflect consistency with the rest of the flow directions, and imposing constraints in order to exclude flow direction selections which are outliers. This way, our existing algorithm has been improved by optimizing the objective function and applying MRF (Boykov et al., 2001) to resolve the topology in sparse datasets without having to aggregate multiple image triplets to resolve the flow directions (see examples from cellular biology in Vid. 1 as well as Vid. 2, which demonstrates the ability of IFTA to resolve point correspondence in a complex rotational motion). For additional examples of tracking results obtained with IFTA, please see (Matov, 2024b; Matov, 2024e; Matov, 2024f; Matov, 2025).

## REAL-TIME MOTION TRACKING

### Feature detection

To detect features in road and human traffic images, we utilized the detector of SIFT (Lowe, 1999; Lowe, 2004), which is an object recognition system has been developed that uses local image features that are invariant to image scaling, translation, and rotation, and partially invariant to illumination changes and affine or 3D projection. The features in SIFT share similar properties with neurons in the inferior temporal cortex that are used for object recognition in our vision. The features are efficiently detected through a staged filtering approach that identifies stable points in scale space (Lindeberg, 1993; Lindeberg, 1998), that is, at different scales of observation. Relevant algorithms on curve estimation, feature statistical significance, bivariate density estimation, clustering, scale-space theory, and edge detection with scale space, respectively, can be found in (Chaudhuri and Marron, 2000; Godtliebsen et al., 1999; Godtliebsen et al., 2002a; Godtliebsen et al., 2002b; Lindeberg, 1994; Perona and Malik, 1990).

The SIFT algorithm generates versions of the image by applying a cascade of DoG band-pass filters for different standard deviations ($\sigma_1$), or scales, with a ratio between the scales of the two distributions ($\sigma_2/\sigma_1$) of 1.4. The detections that persist across scales are selected as image features (keypoints). Additional detectors that will have utility for the analysis of live cell images are SURF (Bay et al., 2008) and the ORB detector (Rublee et al., 2011).

**Closed-circuit television (CCTV)**

Video cameras monitoring the activity of people around sports venues are commonplace in cities worldwide. At sports games, where crowds of tens of thousands gather, such monitoring is important for safety and security purposes. It is also challenging to automate (Rodriguez et al., 2011). Human operators are generally employed for the task, but even the most vigilant individuals may fail to see important information that could ultimately signal the onset of a potentially dangerous situation. We aim at developing a system that can reliably analyze the behavior of up to 200,000 people and vehicles located inside and in the surroundings of a large stadium. Our goal is to create a fully automated accident aversion system, which detects anomalous behaviors in real time as events are progressing. One image CCTV is about 20 Mb or 2.5 MB in size and with 20 frames acquired per second, the bit rate of a single CCTV camera system is roughly 400 Mb/sec or 50 MB/sec.

### Tracking vehicles at crossroads

We tested our algorithm on videos of road traffic crossing at an intersection. Figure 6 shows vectors indicating the features tracked in each of the two directions. Red vectors show the displacement toward the lower-right corner of the image. The figure legend at the upper-left corner of the image indicates that the red flow consists of less features than the light-blue flow and also that it exhibits a low level of angular variation (Fig. 6A), that is, the vehicles within this flow move more or less in the same direction. Conversely, the light-blue flow consists of many vectors and they move in a wide range of directions. This is the result of two lanes (from the bottom of the image and from the right side) merging as the vehicle exit the intersection to the left.

Observe how the speed of the vehicles is reduced (shown with shorter light-blue vectors) as they merge in one lane. Overall, the fastest speeds for both the red and the light-blue flows are measured (and indicated visually as longer speed vectors) in the middle of the image, while the vehicles are crossing the intersection. Conversely, while the vehicles are beginning to cross, as it is with those at the right side of the image, the speeds are low and the vectors are short. The lower image panel (Fig. 6B) is of a time-lapse image later in the sequence and as the vehicles in the red flow are entering the single lane at the lower-right corner of the image, the direction of motion is becoming more homogeneous (see the figure legend at the upper-left corner of the image and compare to the red flow in Fig. 6A).

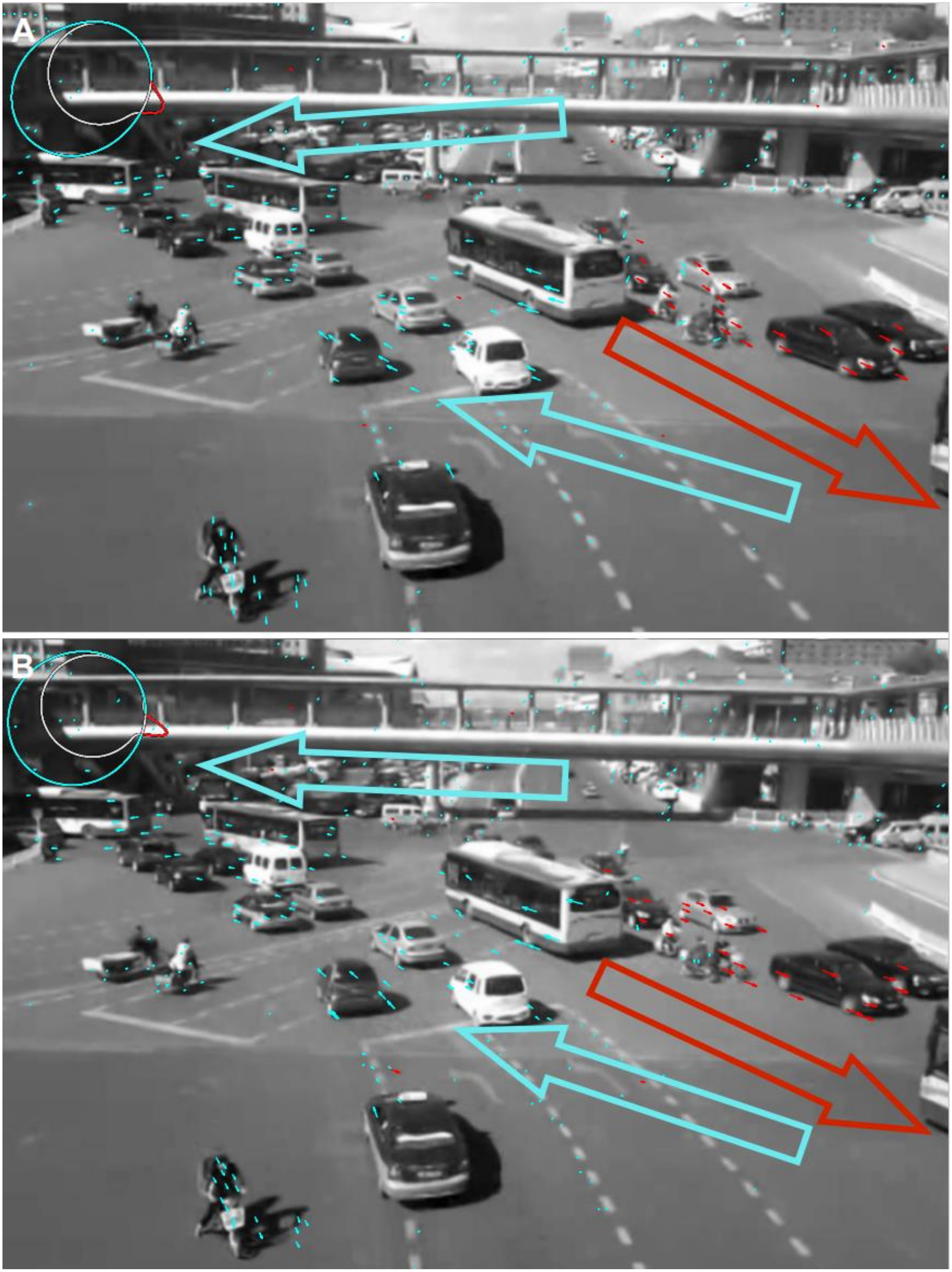

A

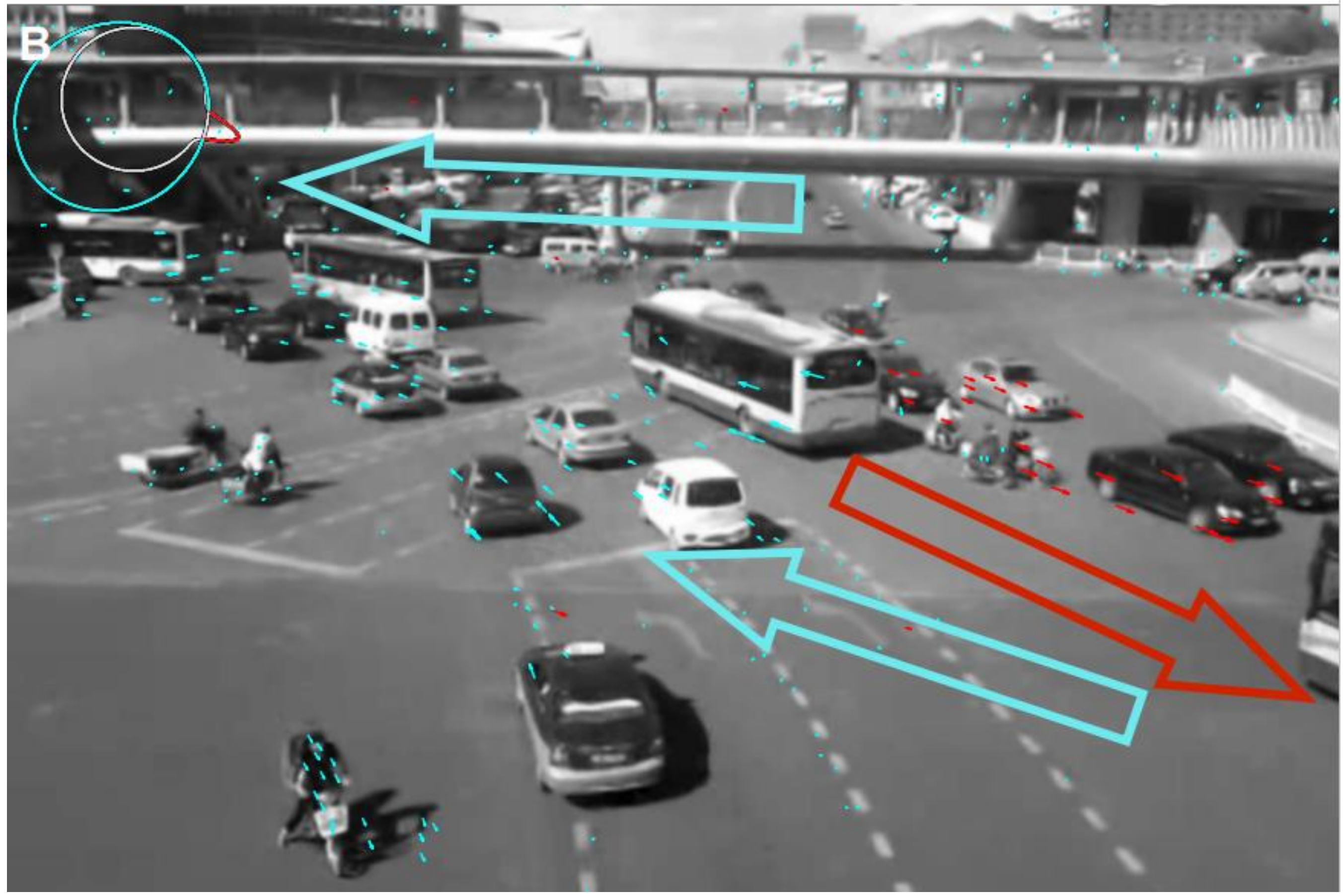

**Figure 6. Tracking road traffic at an intersection.** Two anti-parallel flows are tracked in these images. The main directions are shown as Gaussian distributions at the upper left corner of the image in color-coding. The number of vectors in each flow are reflected in the height of the distributions and the width represents the standard deviation of the direction of motion in each flow. The red flow (vehicles crossing the intersection in the lane toward the lower-right corner of the image) is the fastest with the least standard deviation of the direction of motion. The light-blue flow represents vehicles that merge in the lane on the left side of the image from the lane at the bottom and the lane coming in from the right. (A) and (B) show two of the time-lapse images in the video.

We define the motion in these examples as non-Markovian, that is, the next positions in the trajectories depend not only on the present state, but on the past states as well. The complete trajectories can be computed using a Markovian approximation (Lindblad, 1976; Manzano, 2020). However, it is fair to assume that there is memory in the trajectories, which means that a Markovian assumption might not be applicable. This is due to strong environment couplings, correlations, and entanglement in the motion. A powerful tool for dealing with such systems is provided by the Nakajima-Zwanzig equation (Nakajima, 1958; Zwanzig, 1960). This projection technique has manifested its efficiency at the construction of the generalized master equation and investigation of non-Markovian dynamics.

**Tracking five overlapping flows**

In a large portion of the time-lapse images from the video shown in Fig. 6, we found more than two flows. To analyze a higher number of overlapping flows at the intersection, we performed tracking tests with up to five flows (Fig. 7). Figure 7A shows vectors indicating the features tracked in each of the five directions. Purple vectors show the displacement toward the lower-left corner of the image. The figure legend at the upper-left corner of the image indicates that the purple show consists of a relatively high number of features coming in from the right side of the image and also that it exhibits a low level of angular variation. Conversely, the yellow flow consists of features that represent vehicles coming in from the left side of the image. A few red vectors represent motion from the bottom of the image toward the top of the image. Green vectors track vehicles moving from the top of the image toward the bottom, seen at the intersection under the bridge. Blue vectors track the vehicle moving from the top of the image and turning right.

Further, we analyzed pedestrian (or foot) traffic at an escalator underpass (Fig. 7B) and detected five flows. The purple flow tracked people walking to the left side area of the escalator. The green flow tracked people walking to the right side area of the escalator. The blue flow tracked people taking the wider side of the stairs downwards. As some of the people were turning to the right while leaving the stairs at the wide areas at the bottom of the image, their tracks were grouped with the green flow. The red flow tracked the people taking the escalator upwards. As the upwards escalator is narrow and the people are standing during the motion, this flow exhibited the fastest motion with the least standard deviation of angular orientation (see vector length and the figure legend in the upper-left corner of the image, indicating a narrow distribution of the red angles). The yellow vectors tracked the people who turned to the right to walk into the tunnel as they were leaving the escalator at the top.

Overall, these examples demonstrate the applicability of our algorithm to identify and resolve multiple cohorts of pedestrians and cars in dense traffic scenarios without the need of any user-defined parameters. As shown in the example of anti-parallel flows a biological cell in Fig. 2 (see also the examples in (Matov, 2024b)), IFTA can successfully track up to nine flows when they overlap and are interdigitated. This unique feature of our approach allows us to identify (or detect) different types of urgent situations within only three frames of a streaming video, that is, a fraction of a second after they occur.

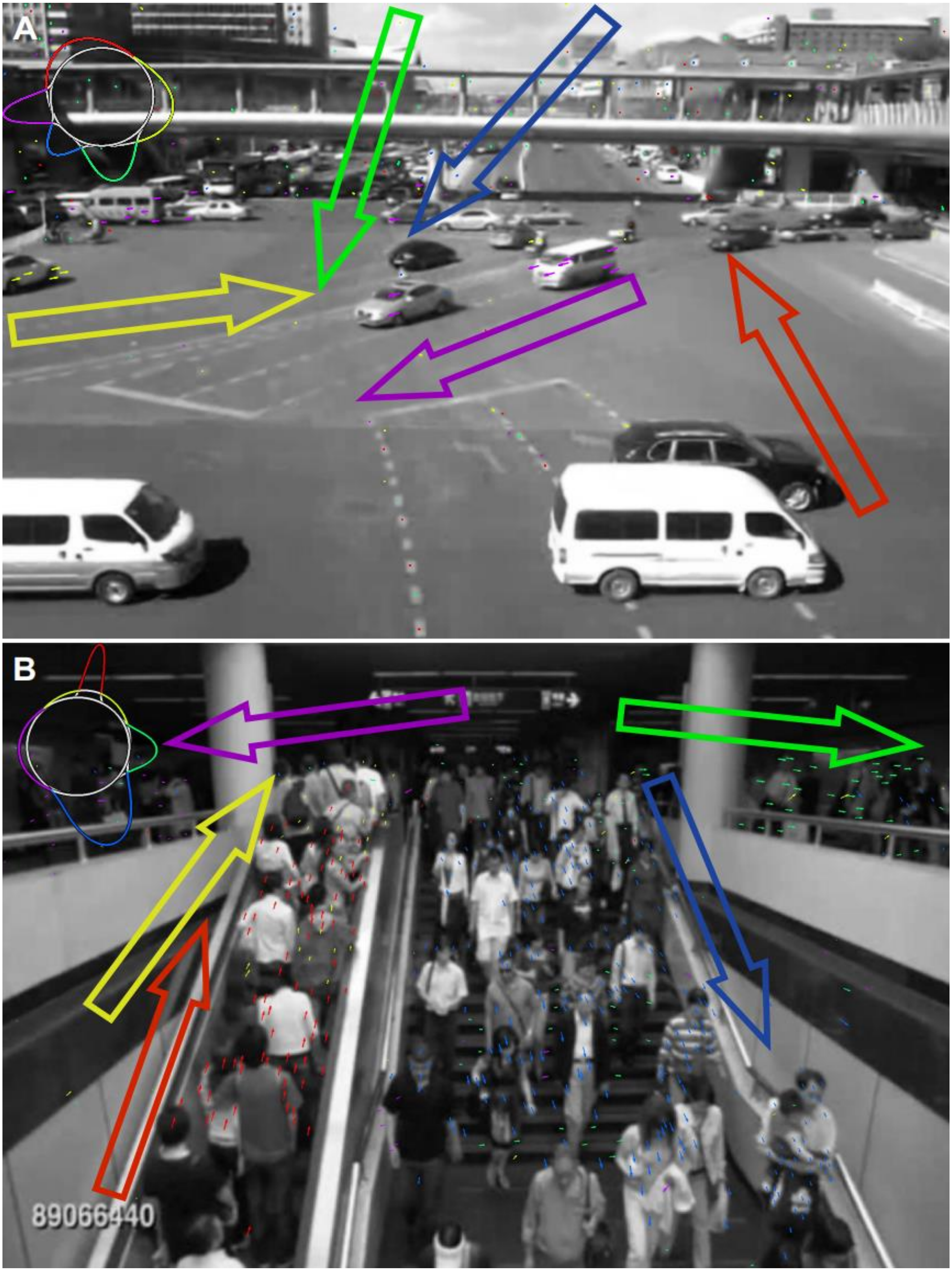

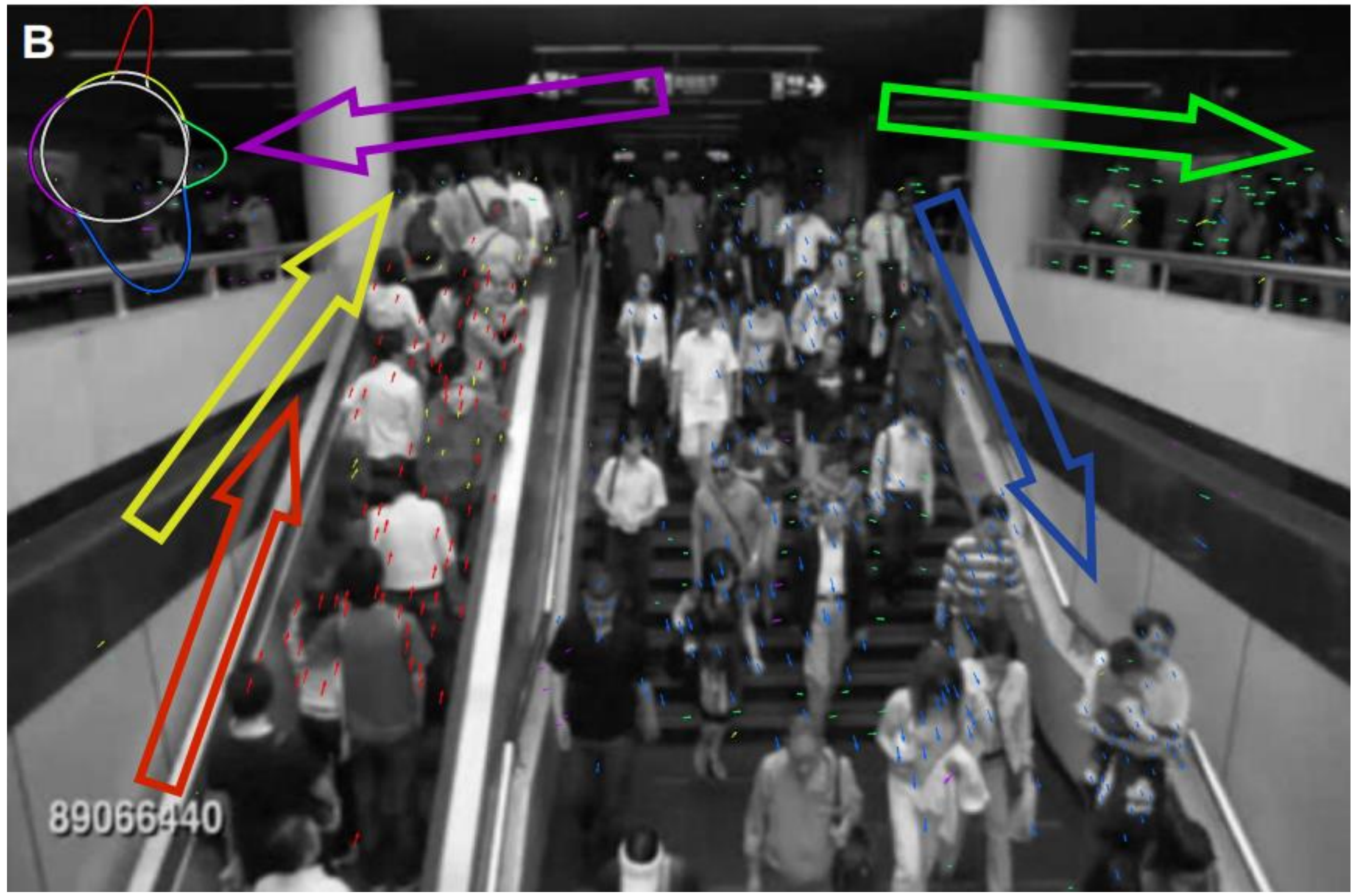

**Figure 7. Tracking five flows of vehicles and people.** Five overlapped flows are tracked in these images. The main directions are shown as Gaussian distributions at the upper left corner of the image in color-coding. The number of vectors in each flow are reflected in the height of the distributions and the width represents the standard deviation of the direction of motion in each flow. (A) Tracking road traffic at an intersection (the same sequence as in Fig. 6). The purple flow (vehicles crossing the intersection from right to left in one lane) is the fastest with the least standard deviation of the angular orientation. The yellow flow tracks the vehicles that move in the opposite direction. The red flow tracks vehicles moving from the lower side of the image toward the upper side. The green flow tracks vehicles moving from the upper side of the image toward the lower side. The blue flow tracks vehicles moving from the upper side of the image toward the lower side, but are turning right (that is, toward the left side of the image). (B) The green flow tracks people walking toward the right side of the escalator area. The purple flow tracks people walking toward the left side of the escalator area. The blue flow tracks people taking the escalator downwards. The red flow tracks people taking the escalator upwards. The yellow flow tracks people walking to the right at the upper end of the escalator.

## Real-time Lucas-Kanade tracking

Our preliminary software, when run on a MacBook with 8 GB of RAM, was able to display 16 frames per second of a synthetic MP4 video of 4.4 MB when no additional graphics were overlaid, that is, using raw image data, and just over 4 frames per second with overlaid flow vectors and other statistical outputs. Our implementation uses robotics computer vision libraries (Abadi et al., 2016; Bradski, 2000; Paszke et al., 2019) and allows us to perform real-time analysis on-the-fly (Matov, 2024g) (see Vid. 3 for an analysis example in a video in high resolution) and a 2D tracking strategy in which we utilize three steps performed by well-established algorithms implemented as a real-time library in OpenCV (Bradski, 2000; Pulli et al., 2012).

In brief, images making up the video are pre-processed or de-noised (Fig. 8) by a specialized background subtraction method (Zivkovic, 2004; Zivkovic, 2006). The algorithm aggregates statistical representation of the intensity profile for each pixel over time during live camera operation by the use of Gaussian (de Moivre, 1718) mixture models (Duda and Hart, 1973) in which the number of Gaussian function components is recursively updated with every new image. Lagrange multipliers (Lagrange, 1788) are introduced in the estimation of the maximum likelihood (Fisher, 1912) to determine the number of components. The influence of previous images in computing the distributions for each pixel is exponentially decaying with time. For multinomial distributions, a conjugate prior is utilized based on a Dirichlet distribution (Dirichlet, 1839) related to the minimum message length criterion (Wallace and Boulton, 1968).

Next, a watershed-based algorithm was used to select the features for tracking (Shi and Tomasi, 1994). The algorithm monitors the quality of the features during tracking by estimating feature dissimilarity between the first and current frames. The section procedure evaluates feature texturedness as it relates to the tracking accuracy. To optimize the tracking accuracy, metrics for two motion models for affine image changes (linear warping) and translation, are computed by Newton-Raphson (Wallis, 1685) minimization. Translation is the better metric when the inter-frame camera translation is small, while affine changes are important for determining dissimilarity between distant frames. The method successfully detects feature occlusions and disocclusions, and selects features based on a threshold for the smaller eigenvalue of the deformation and displacement matrix.

Lastly, the Lucas-Kanade optical flow algorithm, which has been cited some 20,000 times, was used for motion tracking (Lucas and Kanade, 1981). It is an image registration technique that utilizes spatial Gauss-Newton (Gauss, 1809) gradient descent non-linear optimization to direct the search for the feature that yields the best match. While a brute force approach would require quadratic complexity related to the dimensions of the feature, this method is linear in the number of pixels examined in the feature. An initial estimation of the disparity vector is updated via a Newton-Raphson iteration until convergence. The optical flow regression is done via computation of least squares (Legendre, 1805) with pixel weights based on a Gaussian function at the central pixel of the feature. The approach can handle feature rotation, scaling, and shearing.

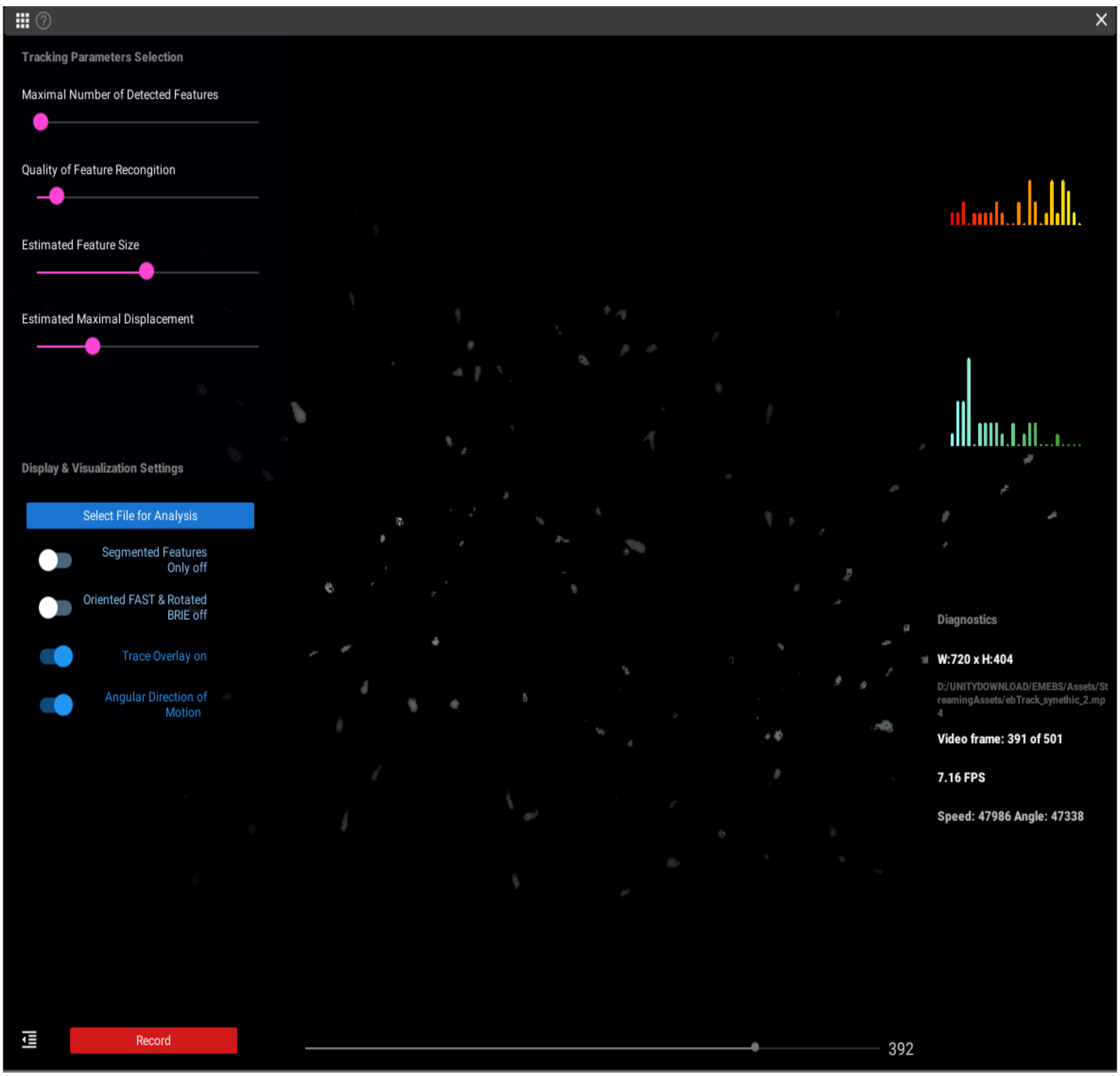

**Figure 8. Real-time segmentation of bright markers.** Each image of the video is pre-processed or de-noised by a specialized background subtraction method. Real-time information on the frames per second analyzed, the average values for the speeds and the angular vector orientations are displayed in the lower right corner of the screen. On the left side of the screen, there are slider buttons in the upper left corner, which allow to set the (i) the upper limit for the number of detected features based on the *a priori* knowledge of the nature of the motion in the analyzed sample, (ii) the level of statistical significance for the feature selection step, that is, the level of feature detection stringency, (iii) the minimum distance between features, which is another parameter selection done based on *a priori* knowledge of the type of sample analyzed, and (iv) a cut-off for the feature search radius, which limits the maximal allowed displacement; this is another parameter, which is selected based on *a priori* knowledge of the sample. By providing sample-specific input to the tracking module, the parameters selection allows to limit the computational complexity, to minimize the tracking errors, and to deliver the fastest analysis results. The blue buttons in the lower-left corner of the screen allow to change various aspects of the screen display in terms of showing segmentation or tracking results, single-segment tracks (between just two frames) or the aggregated trajectories and, as described above, the vector color-coding (angles in red/yellow vs. speeds in different shades of green).

Based on benchmarking with synthetic data, our algorithm had >99% detection accuracy and 98% tracking accuracy. The tracker formed <1% false positive links and about 2% of false negative links, however, these were due on a distance limit we introduced with the search radius. The development environment we used was of the game engine of Unity Technologies (Unity, 2017). We utilized this specific cross-platform engine because it allowed us to compile and test the software on PCs, Macs, smartphones, and smart glasses (see example of tracking cars and boats in Vid. 4). The software suite is suitable for resource-constrained (Khosrow-Pour, 2005), on-the-fly (Happe et al., 2013) computing in microscopes (Edelstein et al., 2010) without internet connectivity (see examples of tracking biological markers in real time by pointing the phone toward the screen in Vid. 5 and Vid. 6). It can be compiled for any operating system and installed on a variety of visualization instruments, like ultrasound and other diagnostic imaging equipment. Our objective has been to present the community with an integrated, easy to use by all, tool for resolving and quantifying a plethora of complex naturally occurring dynamics patterns.

**Generative methods**

It is conceivable that predictions regarding conflict progression can be achieved using generative methods. Generative methods that produce novel samples from high-dimensional data distributions are finding widespread use, for example in speech synthesis. Generative adversarial network (GAN) models (Karras et al., 2017) consist of two convolutional neural networks (CNNs), style-based generator and discriminator, which converge upon reaching Nash equilibrium (Nash, 1950) and can be used to generate synthetic samples. Such sampler posterior distributions can be then used to reduce the complexity and improve the numerical convergence of predicting conflict progression for any new CCTV image using the Bayes formula, given there is available video data to establish transitional states for several conflict and normal situations. Methods for adversarial sample generation in medical data (Zhao et al., 2024) can be adopted as a strategy to augment rare organized-motion instances to correct class imbalance. Our approach may allow the security personnel to attempt inducing changes in reversal (Mullis, 1968) to the deterioration of crowd behavior (Canetti, 1960) occurring during an emerging confrontation, if detected early on. Additionally, our system will be able to detect when small groups of fans are about to confront each other and predict the place of the clash by calculating the speed and direction of motion of the opposing groups (or cohorts) of people.

**Convolutional neural networks**

Action recognition algorithms can deliver 3D image segmentation (Hou et al., 2019; Sun et al., 2024) and feature metrics as the basis for predicting crowd behavior. To simultaneously optimize both the segmentation and assignment steps, we will utilize a 3D CNN (Hou et al., 2019) using separable convolution with dilation, which is suitable for real-time analysis. Adding dilation rates will allow us to capture multi-scale feature representations without significantly increasing the computation cost of 3D convolution. We will also add the utilization of transformer networks (Horiuchi et al., 2024; Ren et al., 2024) for the classification of motion patterns and anomalous motion identification in video surveillance. Tracking data images, in the thousands, with tracks of human crowds visualized, can be utilized for behavior classification by training a generative transformer (Ren et al., 2024), or another large language model (LLM) which takes image patterns as input (Horiuchi et al., 2024). Such a computational approach can aid the selection of an optimal response for each conflict situation and each stage of an ongoing/developing situation can be assigned a dangerous or normal/baseline label and outliers in the behavior of every person in a crowd can be analyzed. All already available in the literature and the public domain computational analysis data can be, thus, re-used within a common AI framework with the objective to aid the evaluation of a situation and inform response.

**Large language models**

We propose the utilization of the described here motion analytics to generate embeddings (Vaswani et al., 2017) for a generative transformer network in which the tokens can be derived from few-shots learning classification (Li et al., 2023) of images of pedestrians and vehicles. We will retrain a transformer network with a new set of tokens - with image classification rather than words and with situation outcomes (that is, accidents) rather than sentences of human speech. This will allow training of the network to classify as safe or dangerous each situation and further identify subclasses with different categories of danger. After training a generative transformer network with image classification and motion dynamics analysis datasets, we will be able to model foot and road traffic as well. Modeling of sequential input data as a Markov process (Bremaud, 1999; Matov, 1999) is an intuitive way to utilize LLMs because of the Markovianity of natural languages (Shannon, 1951). Searching for a comprehensive theoretical analysis of the origins of the impressive performance of LLMs, new research has interpreted LLMs as Markov chains operating on a finite state space of sequences and tokens (Zekri et al., 2024). First order Markov chain models have been shown to learn the correct conditional probabilities by a two-layer transformer (Makkuva et al., 2024). Data will be obtained from traffic cameras and derived during routine surveillance of conditions on the road.

Based on such tracking datasets, conflict detection can be performed by utilizing mathematical reasoning based on a large reasoning model (LRM) (Xu et al., 2025) and reinforcement learning (RL), utilized for post-training of LLMs and inference compute (Guo et al., 2025). Reasoning or a thought process consists of a sequence of tokens representing the steps in the reasoning trajectories. A Kullback-Leibrer (KL) divergence (Kullback and Leibler, 1951) constraint can achieve good empirical results on a range of challenging policy learning tasks (Schulman et al., 2015). The addition of a Group Relative Policy Optimization (GRPO) (Shao et al., 2024), a modification of the RL algorithm for Proximal Policy

Optimization (Schulman et al., 2017), reduces the requirements for training data. For each new dangerous situation, GRPO samples a group of outputs from the old policy and then optimizes the policy model by maximizing the objective function, which is the expectation for the average reward of multiple sampled outputs, produced in response to the same question as the baseline. It utilizes a penalty based on computing the Monte Carlo approximations (Kungurtsev et al., 2023) of the KL divergence between the policy model and the reference model.

RL techniques, such as Markov Decision Processes (Kakade and Langford, 2002; Schneider and Wagner, 1957), can provide the basis for video stratification based on policies for video classification in foot as well as road traffic. In particular, fluid intelligence (Cattell, 1943) applications of LRMs that depend on no prior training data can be useful in identifying and detecting new types of conflict situations in real time. Upon validation, our computer vision approach for analysis of foot and vehicle traffic can become a broadly applicable tool for motion analytics based on a LRM.

**Use cases**

Our technology can be used to analyze videos of conflict situations at different types of sports stadiums, for example, for association football (Fig. 9) and American football (Fig. 10). Furthermore, we can envisage additional applications such as using the technology to detect dangerous behaviors at airports (Fig. 11), train station (Fig. 12), political rallies/mass demonstrations (Fig. 13), large retail stores (Fig. 14), music festivals (Fig. 15), and busy resorts (Fig. 16), among a number of other important security applications.

Our efforts are focused on the development of a system that provides the security personnel on-the-fly with an automatically generated alert signal regarding rapid motion of groups of individuals or events of interest in crowded scenes. In addition, the system will offer crowd density estimation using our novel approach and the prediction of overcrowding in front of the gates and parking lots around the sports arenas, which may occur without any intention and by chance, depending on the times of arrival of the spectators (Still, 2000).

Our solution is especially beneficial for optimizing security responses, enhancing operational efficiencies, and enabling data-driven decision-making for event organizers, transportation managers, and urban planners. Potential use cases include (i) sporting events and stadiums – tracking crowd movement during games and events to optimize concession sales, reduce wait times, and enhance overall fan experience while ensuring safety, (ii) transportation and airports – real-time tracking of passenger flows in airports, train stations, and bus terminals to optimize security, reduce wait times, and improve the passenger experience, (iii) smart cities and public safety – monitoring public spaces, transportation hubs, or urban areas to better manage crowd flow, improve safety, and respond to emergencies, (iv) retail and commercial spaces – monitoring foot traffic in malls, shopping centers, or large retail stores to optimize store layouts, marketing campaigns, and staffing, (v) event management – real-time tracking of crowd density at concerts, festivals, sports events, and other mass gatherings to prevent accidents, optimize resource allocation, and enhance customer experience, (vi) tourism and attractions – monitoring visitor flows at tourist attractions, museums, and theme parks to enhance guest experiences, manage crowd control, and optimize staffing during peak seasons, (vii) workplace management – analyzing employee foot traffic in office buildings to optimize space utilization, enhance workplace safety, and improve operational efficiency, and (viii) healthcare facilities – monitoring patient and visitor flows in hospitals and healthcare centers to optimize staff allocation, enhance patient experience, and manage emergency situations.

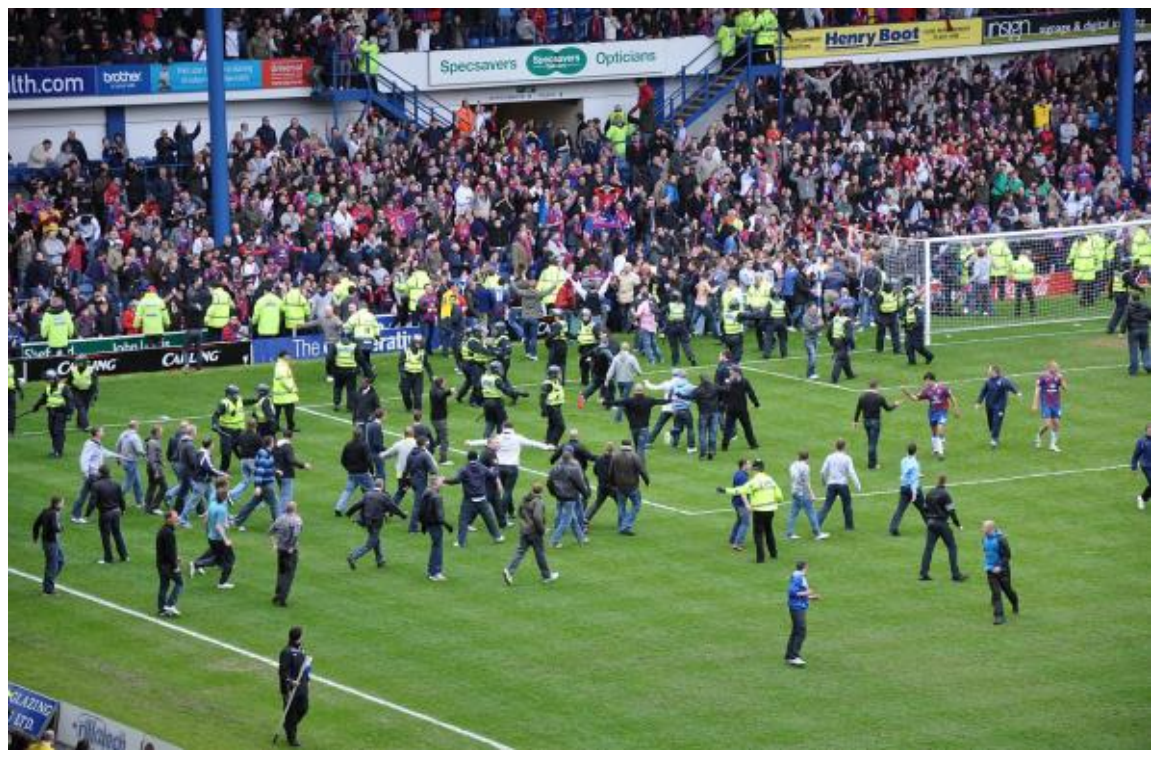

**Figure 9. Fans at an association football stadium.**

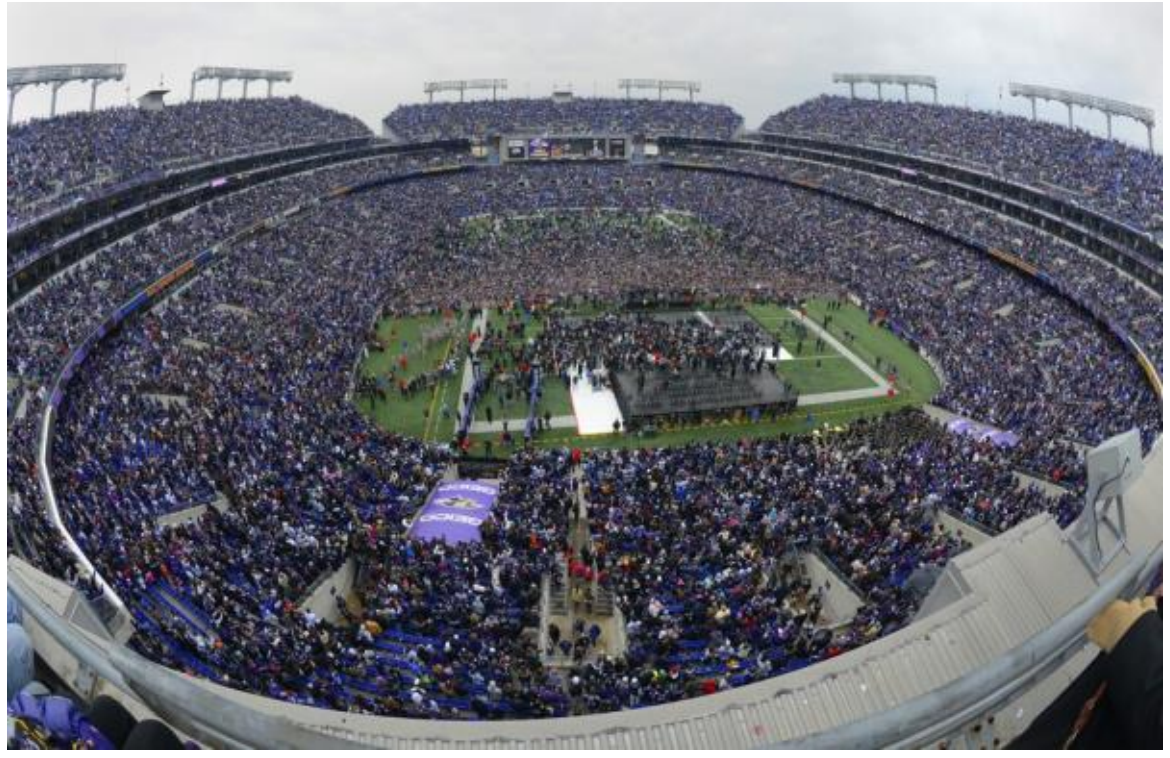

**Figure 10. Fans at an American football stadium.**

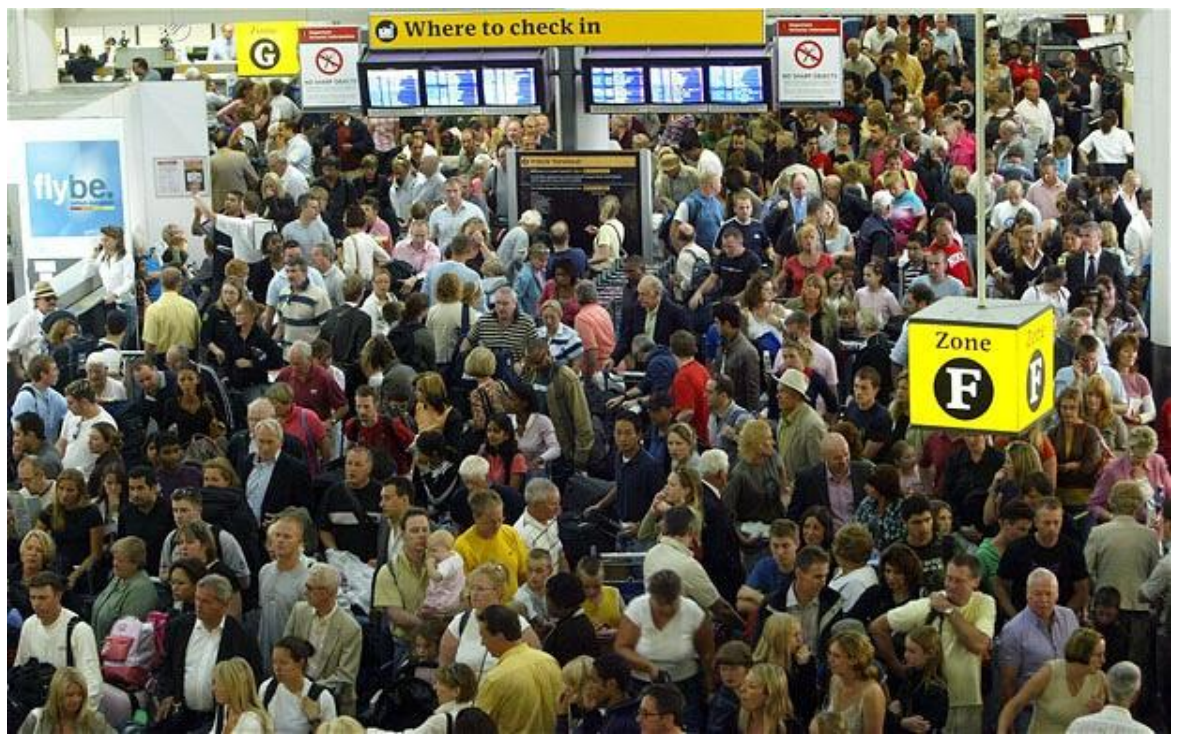

**Figure 11. Passengers at an airport terminal.**

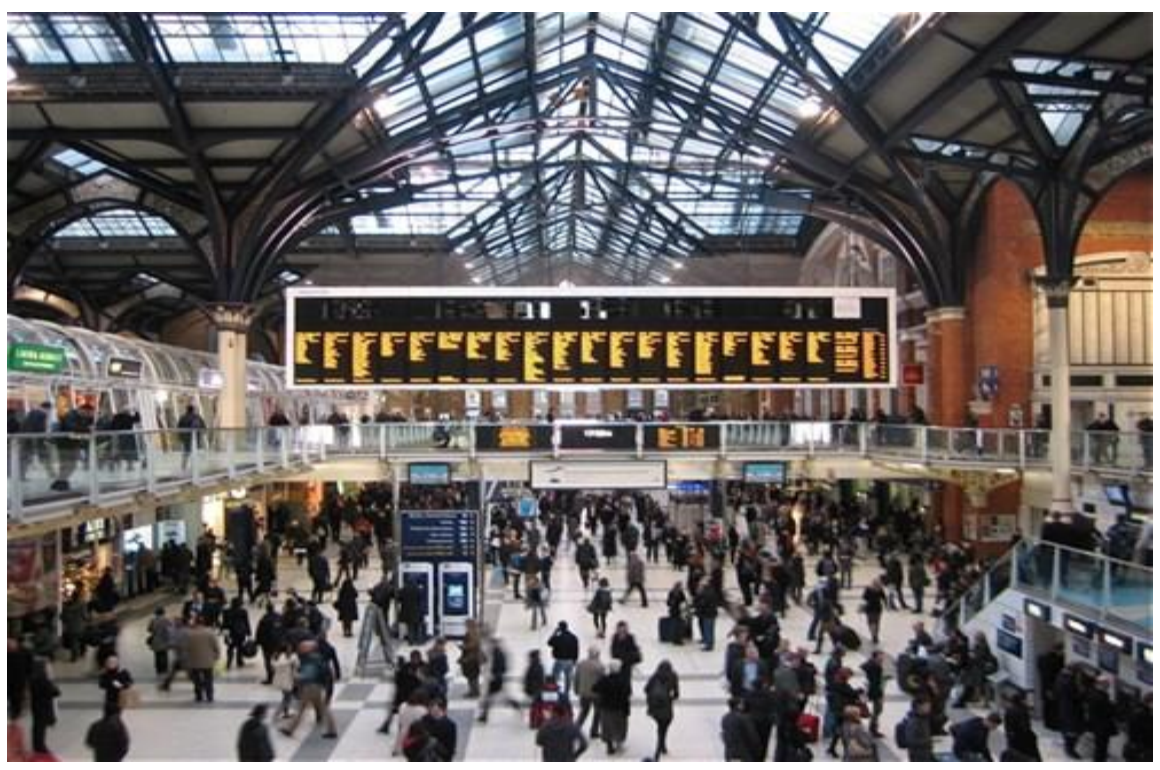

**Figure 12. Passengers at a train station.**

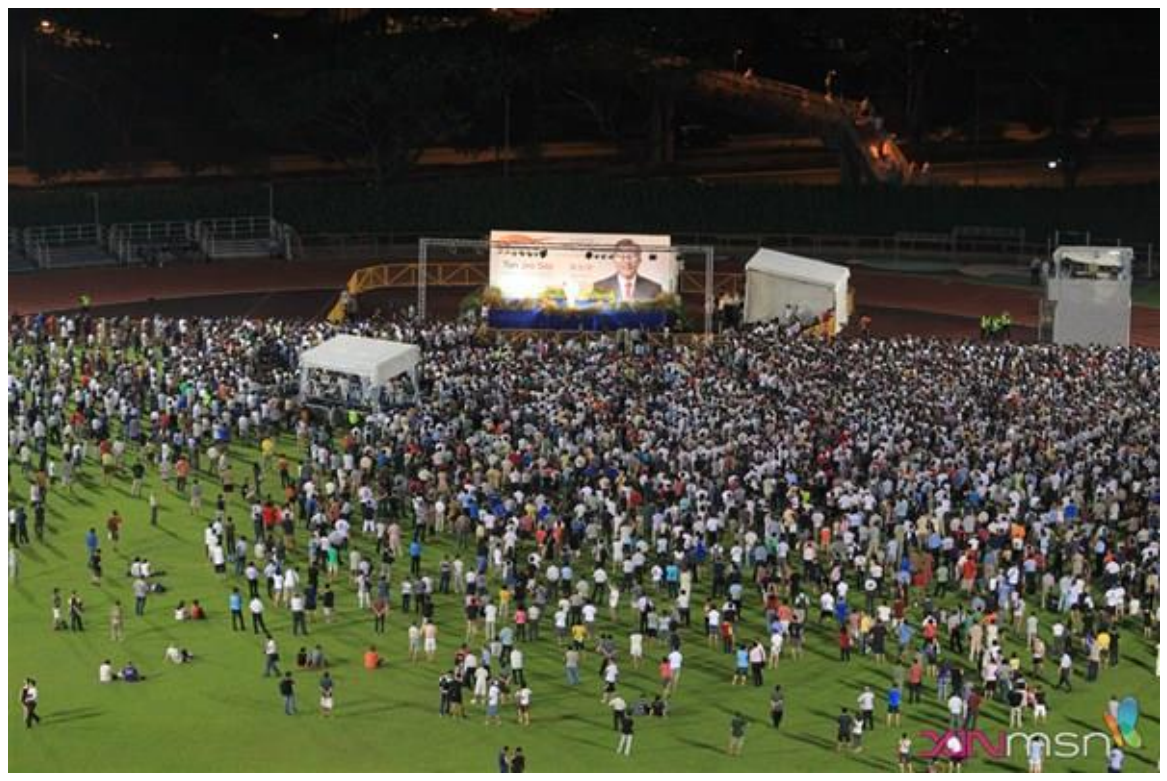

**Figure 13. Spectators at a political rally.**

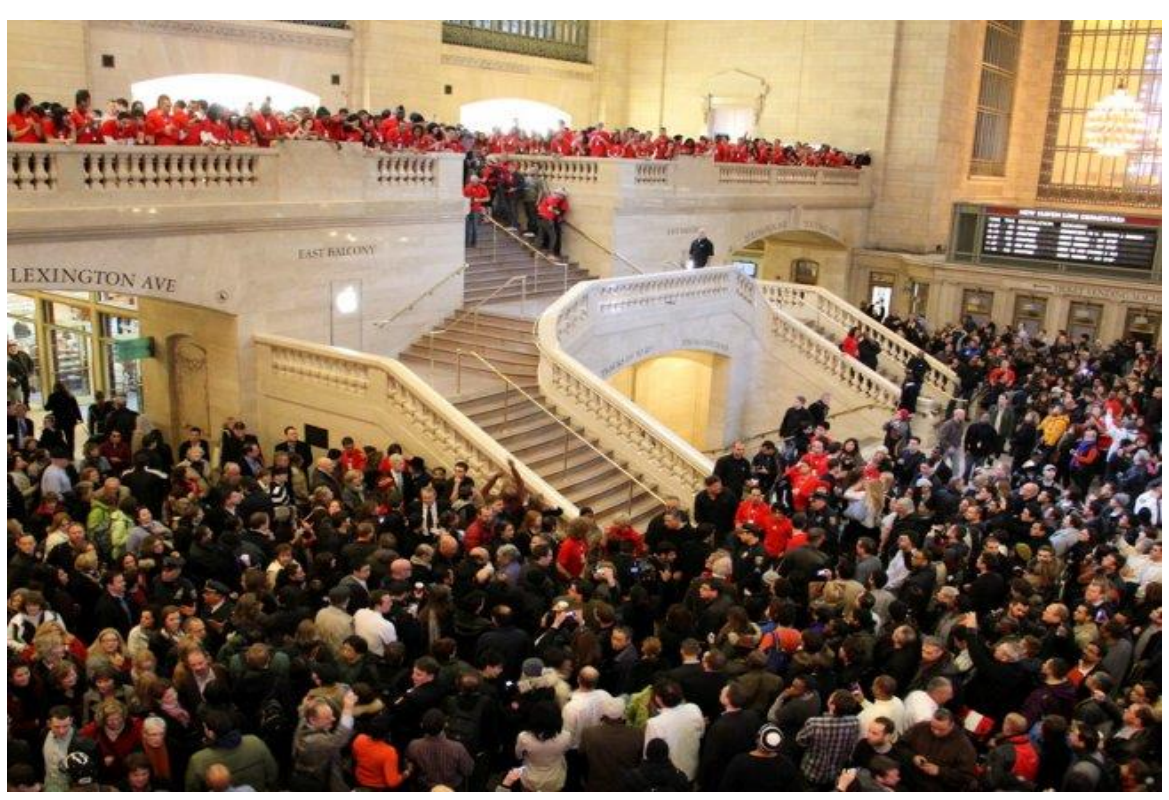

**Figure 14. Customers at an Apple store.**

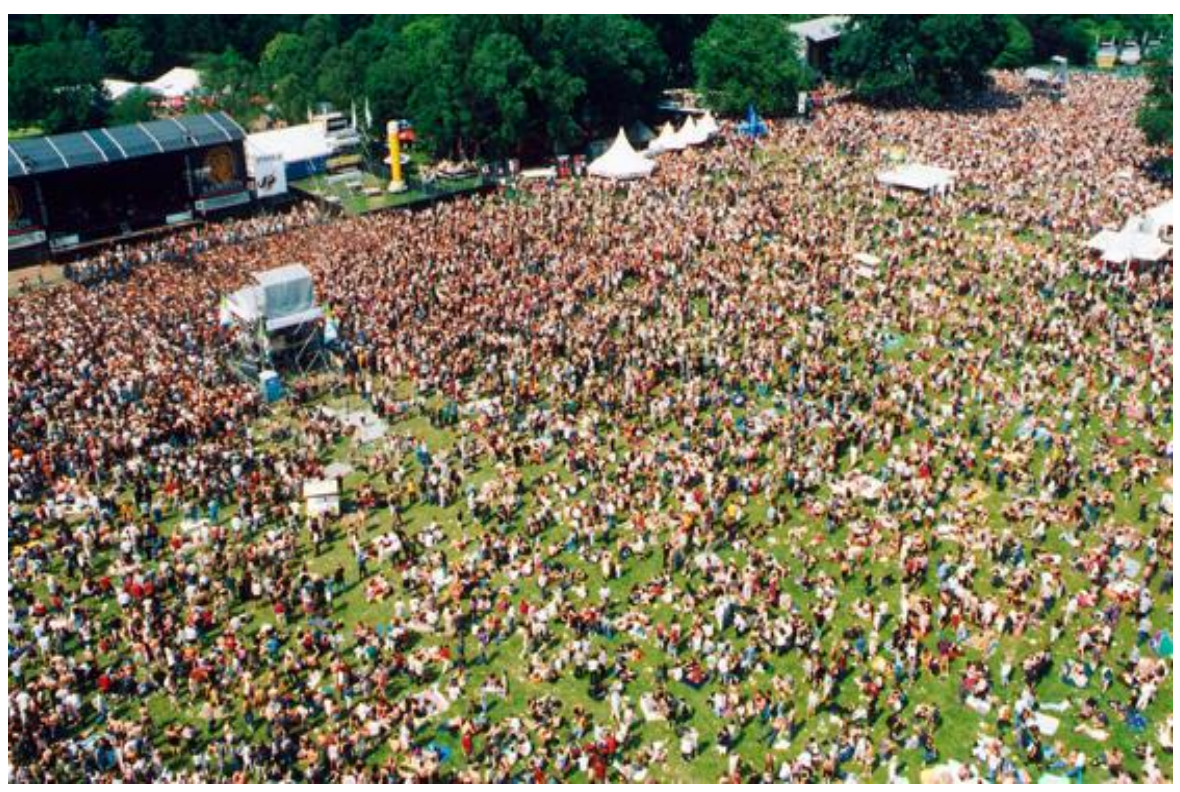

**Figure 15. Spectators at a music festival.**

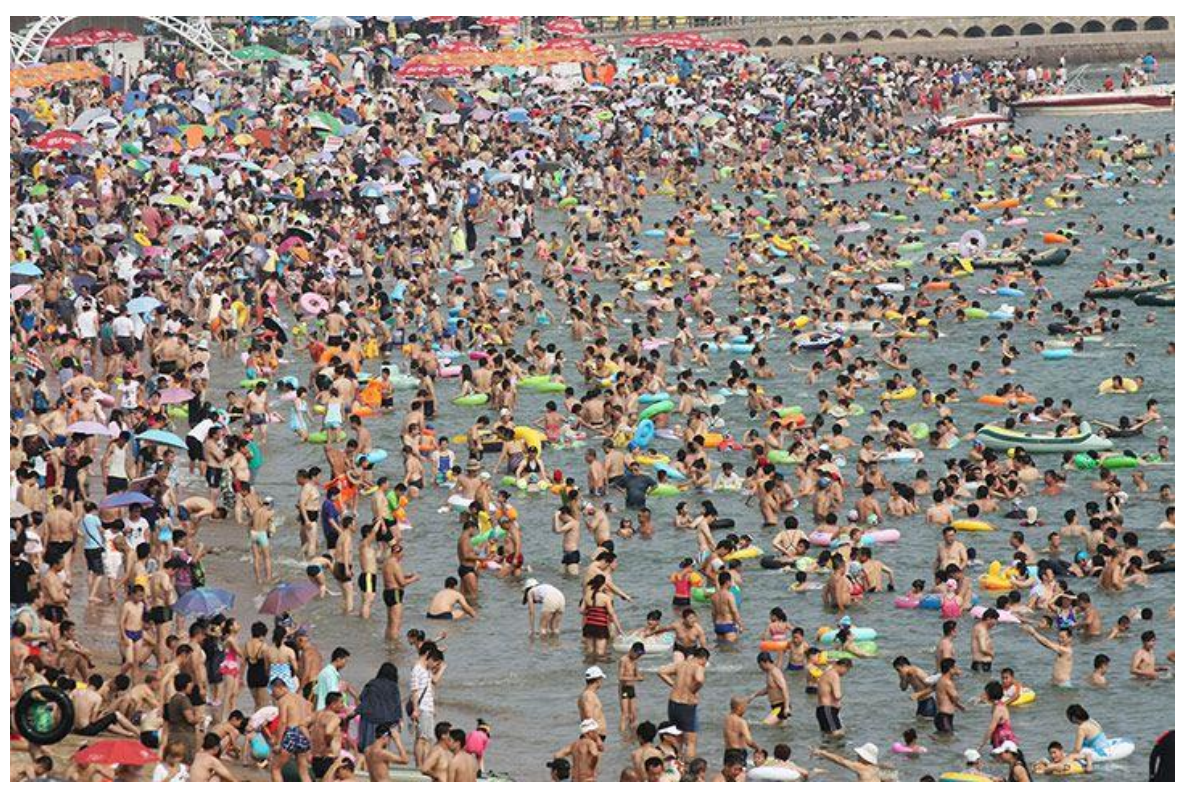

**Figure 16. Tourists at a summer resort beach.**

## DISCUSSION

Crowd management and safety are critical concerns in industries such as retail, event management, transportation, and smart cities. Large gatherings, whether planned or spontaneous, pose challenges including overcrowding, safety hazards, operational inefficiencies, and limited real-time insights into crowd behavior. Events with large crowds are associated with (i) inability to manage real-time crowd dynamics and density, (ii) an increased risk of accidents, stampedes, or panic (Helbing et al., 2000) in overcrowded areas, (iii) an inefficient use of resources such as security personnel, emergency services, or crowd flow management tools, and (iv) an overall lack of detailed insights into foot traffic, behavioral trends, and demographics, which hinders data-driven decision-making for revenue optimization, safety protocols, and operational efficiency. Importantly, our unique approach can instantly identify a group of people (or drones) that suddenly begin moving against a common target, such as an airport gate, or when groups of sports fans are clashing and provide precise spatial information, which will lead to avoiding such incidents.

We will develop Software-as-a-Service (SaaS) applications for various platforms and devices, beyond CCTV camera systems, such as smartphones (iOS / Android) as well as smart glasses (HTC Vive / Microsoft HoloLens) (Matov, 2024f; Matov, 2024g). Furthermore, we envisage using the existing infrastructure of video cameras for acquiring image datasets on-the-fly and deploying an easy-to-use data-driven software system for parsing of significant events by analyzing image sequences taken inside and outside of sports stadiums or other public venues. Other prospective end users are organizers of political rallies, civic and wildlife organizations, security firms, and the military. Local area networks for surveillance can be established based on powerline communication (Matov, 2001a; Matov, 2001b) systems in remote locations, such as the polar areas. We will optimize the performance of the software by implementing a classification method able to distinguish between activities posing a threat and those not posing a threat. In this context, our algorithms can be utilized by video analytics companies, such as LenovoEMC (CN) (Lenovo, 2014) and ObjectVideo (Alarm.com) (ObjectVideo, 1998).

Our technology can be embedded in a new generation of compact police car and fire track as well as school and border cameras, which will provide a detailed foot and vehicle traffic analysis. Our software can be embedded in the hardware of network video recorder companies focused on surveillance use cases, such as DVTel (Teledyne FLIR) (DVTel, 2000), Milestone Systems (Canon, DK) (Systems, 1998), and Genetec (CA) (Genetec Inc., 1997), as well as megapixel camera specialists like AV Costar (Arecont Vision, 2003), Mobotix (Konica Minolta, JP) (Mobotix, 1999), and Axis Communications (SE) (Axis Communications AB, 1984).

To our knowledge, an algorithm able to resolve the topology of dense, overlapped flows is not used in the public space – neither in the US or elsewhere. Our computational algorithms can be integrated in the software of AI analytics companies for smart surveillance, such as Ultralytics (Ultralytics, 2022), Irisity (SE) (Irisity AB, 2006), Intelli-Vision (Nortek Control, 2002), Verkada (Verkada, 2016), Coram (Coram AI, 2021), Eagle Eye Networks (JP) (Eagle Eye Networks, 2018), Gorilla Technology (UK) (Gorilla Technology Group, 2001), and Rhombus (Rhombus Systems, 2016), which perform pedestrian or vehicle traffic analysis.

Further, Beca (NZ) (Beca, 1920) and Bentley Systems with their software Legion (Legion, 1997) are the companies that perform crowd analysis in the crowd modeling (Makinoshima and Oishi, 2022) industry. Crowd Dynamics (UK) (Dynamics, 2010) performs similar crowd modeling as well as transport planning. However, these companies only model the overall flow – in the context of infrastructure planning in civil engineering and the design of venue architecture – and do not offer real-time conflict avoidance methodology. Density (Density, 2014) offers software for people counting in real time to optimize space utilization and assist in occupancy planning.

Several companies offer tracking at airports and road intersections, but these analyses are limited to a small number of people and cannot handle crossing of motion, which is common. CrowdVision (Skyfii, AU) (CrowdVision, 2009) has deployed airport software systems for nearest neighbor tracking (similar to (Cronin and Stumpe, 2014)) and a dominant (single) flow tracking for real-time computer vision to optimize queue management. Xovis (CH) (Xovis AG, 2009) has similarly developed and tested airport software systems for real-time computer vision that perform individual passenger motion tracking in sparse crowds and crowd density estimation – also within public transport vehicles – to limit waiting times.

Our technology can be made available as turn-key software or through a web interface via which additional modules with algorithms to analyze images of specific types of motion in streaming video or other imaging modalities can be added as needed by the SaaS application. Additional modules with specialized algorithms, including video-tagging of spectators (or passengers), to analyze images with highly unusual types of motion and behavior from live online cameras or other imaging devices will be implemented. Our goal is to create a fully automated and efficient real-time accident aversion system, which can be integrated within the technological developments taking place in contemporary urban areas, that is, the so-called smart cities.

## MATERIALS AND METHODS

### Image analysis

All image analysis programs for detection and tracking, and graphical representation discussed were developed in Matlab, C/C++, and C#. The SIFT detection method used is described and validated in (Vedaldi and Fulkerson, 2008), the tracking method IFTA is described and validated in (Matov et al., 2011). The computer code is available for download at: https://www.github.com/amatov/InstantaneousFlowTracker. The CPLEX optimizer requires an ILOG license from IBM.

The methods for real-time feature detection and tracking, and graphical representation of the results were developed in the cross-platform game engine Unity in C# (Matov, 2024g). The computer code is available for download at: https://www.github.com/amatov/DataSetTracker. OpenCV for Unity requires a license from Enox Software.

To test the software, we have uploaded EBTracker.exe and DataSetTracker_v1.1.zip files at the same link. In the zip folder, there are instructions on how to test the tracker. It does not require any installations and starts on double click. A default test video is included with the download and the analysis begins upon clicking the software executable file. The user can change the video for analysis.

**Abbreviations**

CCTV – closed circuit television

CNN – convolutional neural network

DoG – difference of Gaussians

GAN – generative adversarial network

GRPO – group relative policy optimization

IFTA – instantaneous flow tracking algorithms

LLM – large language model

LRM – large reasoning model

MRF – Markov random field

MT – microtubule

OpenCV – open computer vision library

ORB – oriented FAST (features from accelerated segment test) and rotated BRIEF (binary robust independent elementary features)

RL – reinforcement learning

SaaS – software-as-a-service

SIFT – scale-invariant feature transform

SURF – speeded-up robust features

## Data availability statement

The datasets used and/or analyzed during the current study are available from the corresponding author upon reasonable request.

## Conflicts of interest

The author declares no conflicts of interest.

## Funding statement

No funding was received to assist with the preparation of this manuscript.

## Author contributions

A.M. conceived the study, developed software, performed computer vision and statistical analysis, prepared figures and videos, and wrote the manuscript.

## ACKNOWLEDGEMENTS

The images shown in Fig. 6, 7, 9 – 16 were downloaded from the internet. I thank Shayan Modiri and Mubarak Shah for discussions in regard to this work. A preprint of this paper is available at arXiv (Matov, 2024a).

## SUPPLEMENTARY MATERIALS

Video 1 – Time-lapse video of a bipolar mitotic spindle. The signals tracked are diffraction-limited sub-resolution clusters of fluorophores (Matov, 2024e) within the lattice of polymerizing proteins called microtubules, which form the anti-parallel cohorts of features moving in opposite directions during cell division. https://vimeo.com/1059101952/da58c7a145

Video 2 – Tracking in a rotating mitotic spindle. The video shows a dividing breast cancer cell from (Matov, 2024d). The signals tracked are diffraction-limited sub-resolution clusters of fluorophores within the lattice of polymerizing proteins called microtubules, which form the anti-parallel cohorts of features moving in opposite directions during cell division. The flow fields' vectors are color-coded in shades of red and blue to distinguish the two cohorts. https://vimeo.com/999199825/831d5f0dc1

Video 3 – A demo video of the functionality of real-time image analysis software suitable for resource-constrained computing (Matov, 2024g) in high resolution. https://vimeo.com/manage/videos/238144544

Video 4 – Tracking of road traffic in real time through a smart phone app. Real-time tracking of cars and boats from a live feed via a phone app using (Matov, 2024g). The measured speeds are in pixels per frame. https://vimeo.com/1037963282/164219f147

Video 5 – Tracking in real time through a smart phone app. Real-time tracking of fluorescent markers in a time-lapse sequence via a phone app using (Matov, 2024g). This example was acquired during app development. The phone is pointed toward a computer screen displaying a biological sample with moving fluorescent markers. The red markers on the phone screen are the detected and tracked signals. https://vimeo.com/1037964695/9a9d4e0b5d

Video 6 – Tracking in real time through a smart phone app. Real-time tracking of fluorescent markers in a time-lapse sequence via a phone app using (Matov, 2024g). This example was acquired during app calibration. The two sliders on the side of the phone screen allow for fine-tuning of the detection and tracking parameters. The phone is pointed toward a computer screen displaying a biological sample with moving fluorescent markers. The red markers on the phone screen are the detected and tracked signals. The measured speeds are in pixels per frame. https://vimeo.com/1037965812/c270ae4844